# A New Approach for Arabic Handwritten Postal Addresses Recognition

Moncef Charfi, Monji Kherallah, Abdelkarim El Baati, Adel M. Alimi
REGIM: Research Group on Intelligent Machines,
National Engineering School of Sfax, University of Sfax, Tunisia

*Abstract*—In this paper, we propose an automatic analysis system for the Arabic handwriting postal addresses recognition, by using the beta elliptical model. Our system is divided into different steps: analysis, pre-processing and classification. The first operation is the filtering of image. In the second, we remove the border print, stamps and graphics. After locating the address on the envelope, the address segmentation allows the extraction of postal code and city name separately. The pre-processing system and the modeling approach are based on two basic steps. The first step is the extraction of the temporal order in the image of the handwritten trajectory. The second step is based on the use of Beta-Elliptical model for the representation of handwritten script. The recognition system is based on Graph-matching algorithm. Our modeling and recognition approaches were validated by using the postal code and city names extracted from the Tunisian postal envelopes data. The recognition rate obtained is about 98%.

## I. INTRODUCTION

For several years, on-line and off-line hand-writing recognition has been considered [8], [24]. The postal automation, bank checks identification, automatic processing of administrative files and cultural patrimony heritage are direct applications of the optical character recognition, which present difficult problems due to the presence of handwritten manuscripts in such documents.

Postal automation has constituted a potential application of character recognition and a real driving challenge feeding research in such domain. Everywhere in the industrialized world, the postal services have financed and continue to finance lots of works in such complex domain. Today, in the western countries, millions of postal objects essentially made up of letters and parcels, are collected in mail sorting centres for their redistribution. These systems recognize the postal address, and print a bar code on the envelope. Letter forwarding is based on the reading of the addresses' bar code, thus making it possible to analyze 50000 letters per hour [32],[35].

On the other hand, the Arab and Eastern countries are somewhat behind in this domain. To take up the challenge, we have considered the problem of the automatic process of Arabic postal addresses in Tunisia and Arab countries.

Our paper is formulated as follows: section 2 presents a state of the art of handwritten character recognition, and the problems encountered in the process of Arabic postal addresses. Paragraph 3, describes the pre-processing steps, and paragraph 4 gives the temporal order reconstruction. In section 5 we present the beta elliptical approach for handwriting modelling. Section 6 is devoted to the recognition and some experimental works discussion.

## II. STATE OF THE ART AND PROBLEMATIC

The postal automation systems are generally based on printed or off line handwriting recognition. Printed characters are now well recognized, but handwritten character recognition remains very difficult, on account of its very great variability. We find in the literature several descriptions of such systems. Wada [33] proposed a total system of automatic analysis of addresses. Gilloux presented a postal addresses recognition system [13]. Heute proposed a system of postal automation as a potential application for recognition digits and characters manuscripts [15], [16]. In such systems, the envelopes submitted to recognition are ordinary (handwritten or printed) and present the greater part of the difficulties that a recognition system has to face: address location, separation printed/manuscript, segmentation in lines, words and characters, inclination correction, etc. In [3], text/graphic segmentation which is based on the detection of the geometric model detection of the related components on an envelope, and on graph discrimination such as stamps and logos, is done by computing pixel density. Menoti and all, present a segmentation algorithm based on feature selection in wavelet space. The aim is to automatically separate in postal envelopes the regions related to background, stamps, rubber stamps, and the address blocks [26], [34], [36].

The techniques developed for OCR systems are generally based on the neural networks and Markovian approaches [31]. In fact, several recognition systems in the literature are based on Hidden Markov Models [20], [37]. Lee and all [21] have developed a new hybrid approach to the verification of handwritten addresses in Singapore. The hybrid verification system seeks to reduce the error rate by the correlation of the extracted postcode features set recognized words from the original handwritten address. Novel use of syntactic features extracted from words has resulted in a significant reduction in the error rate while keeping the recognition rate high [30].

For the degraded character recognition, Likforman-Sulem and Sigelle proposed in 2008 models based on the formalism of Bayesian networks to re-introduce the notion of spatial context [22]. Bayesian network is a probabilistic graphical model in which links are introduced between distant observations. Similarly, models based on recurrent neural networks extend classical neural networks by introducing the notion of context by bidirectional dependencies. In particular the model LSTM (Long Short Term Memory) has been recently applied to the recognition of cursive online [14].

Gaceb et all, present a new approach for address block location based on pyramidal data organization and on a hierarchical graph colouring for classification process. This new approach permits to guarantee a good coherence between different modules and to reduce the computation time and the rejection rate, and gives satisfying rate of 98% of good location [11], [12].

Liu et all, proposes an approach to retrieve envelope images from a large image database, by graph matching. The attributes of nodes and edges in the graph are described by characteristics of the envelope image, and the results of experiments tests using this approach are promising [23], [29].

In Arabic optical character recognition (AOCR), Lorigo presented a study focusing on the off-line Arabic handwriting recognition. She proved that Arabic writing presents important technical challenges which have to take up [2], [17], [24], [28] and [31].

In the case of on-line handwriting, the dynamic information, such as the temporal order and pen speed, make a performance for handwriting modelling. In this context, the reconstruction of the temporal order of off-line handwriting improves the performance of recognition systems [1] [7].

In the case of the printed and scanned documents, the results are very interesting if the images are high quality. However, these results decrease quickly if the image contains noise as stain, background, etc. Based on the complementarities existing between classifiers (Multi-Layer Perceptron (MLP) [5], Hidden Markov Model (HMM) [10], Genetic Algorithm (GA) [19], Fuzzy Logic [6], etc.),

The majority of classifiers meet a major problem which lies in the variability of the vector features size. In literature, three approaches are commonly used to manage the problems of dimensionality. These approaches are: Genetic Algorithms, Dynamic Programming and Graph Matching. In our work, the method of graph matching is appropriately used to cover the problem of dimensionality of feature vector, as was done by Namboodiri and Jain in the case of Graph Matching use for Arabic word recognition, and by Rokbani and al., in the case of the Genetic Algorithm and Graph Matching combination for on-line Arabic word recognition [25], [28].

The researcher in the Arabic postal envelopes processing encounters many difficulties: extraction of addressee address, extraction of postal code, segmentation of postal code in digits and variability of writer.

In the absence of standard format of envelopes, locating the address becomes difficult and depends on locating all objects constituting the address [4], [33], [35]: analysis, location and description of interest zone. Besides, it's noticeable that the handwritten address presents some difficulties and problems, like the inclination of the baseline of the handwritten address, and the variable position of the address on the envelope, which requires correction and particular processing.

Addressee's address, deriving from the extraction, contains the numerical and alphabetic data. A stage of discrimination (digit/letter) is therefore necessary. The position of the postal code in the address is variable according to the writer.

Analysis of the postal code depends on its decomposition into digits. But, this decomposition is not always possible because several types of connections exist between the handwritten digits:

- Simple connections, where only one link exists between two strokes of writing;
- Multiple connections, where at least three strokes of writings exist on at least one connection [15], [24].

The pattern of handwritten writing is very variable. It translates the style of writing, the mood and the writer's personality, which makes it difficult to characterize (see figure 1):

- The chosen value of image resolution is important because it conditions the recognition system. It is very important to choose a high resolution ($\geq$ 300 ppi) to keep the maximum of information on the image.
- Noise and irregularities in the tracing are at the origin of the disconnection of some features.
- The distortions are the dilations, narrowing and other local variations in the writing.
- The variability of the style, that is the use of different forms to represent the same character, such as the flowery style or the slope and, in a general manner, everything that characterizes the writer's habits.
- The translation of the whole character or a part of its components.
- The dissymmetry that leads to a closure and a complete closing of the digit.
- The bad adjusting of the digits causing abrupt interruptions of the lines.

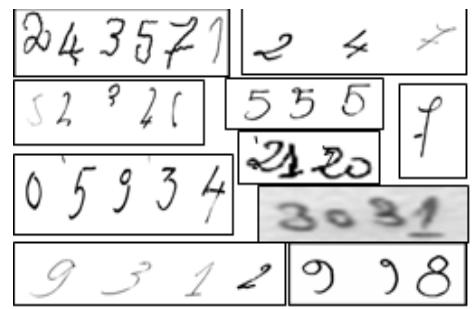

Figure 1. General problems in digits recognition

- The slope of the digits and the baseline: different directions can be seen on the same page because the changes of paper's orientations during writing.
- A handwritten digit cannot represent the same features as the set of other digits of same family, and therefore generates confusions [14] (see figure 1).

### III. PRE-PROCESSING STEPS

The pre-processing steps consists on: filtering of the image, location and suppression of printed borders, location and suppression of stamps and graphics, address location, segmentation of address in lines, segmentation of the last two lines, and segmentation into connected components [5].

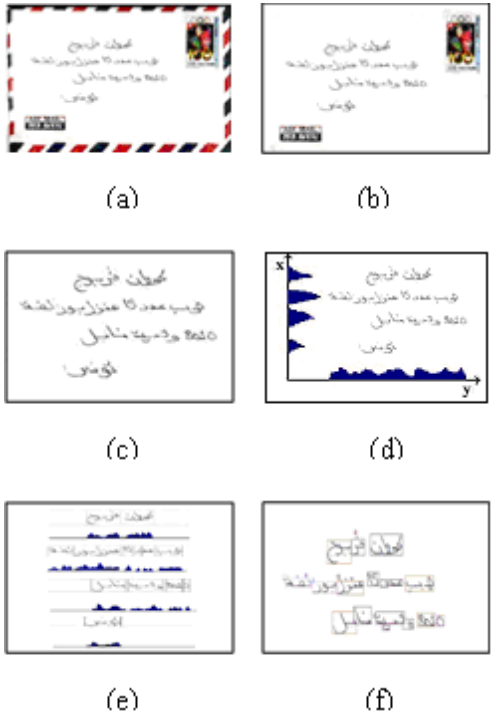

Figure 2. Pre-processing steps

(a) Image of envelope,
(b) Location and suppression of printed borders,
(c) Suppression of stamps and graphics,
(d) Address location,
(e) Address segmentation in lines by horizontal projection and in words by vertical projection,
(f) Words of address segmented.

In figure 2, we regrouped all illustrations of the main pre-processing steps.

The final step is the discrimination postal code/city name that is based on the test of regularity of connected components and test of eccentricity.

### IV. TEMPORAL ORDER RECONSTRUCTION

The handwritten of a word is the stroke constituted by a set of curved lines. Every line has a starting point and an ending point. Before the temporal order reconstruction, we firstly proceed to skeleton segmentation. Word image, representing the city name, goes through three stages of pre-processing: binarization, filtering, skeletisation and elimination of the diacritic signs, as the points above and below the words and the vowel signs, (see figure 3). Three types of characteristic points will be extracted from the skeleton of the tracing [7]:

- The end stroke point: this is the black pixel that possesses only one neighbour of the same type.
- The branching point: this is the black pixel that possesses three neighbours of the same type.
- The crossing point: this is the black pixel that possesses four neighbours of the same type.

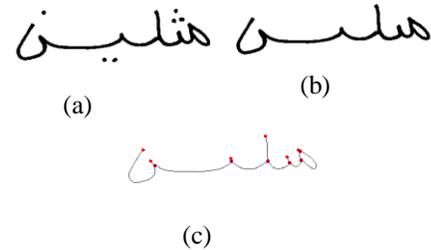

Figure 3. Pre-processing steps

(a) Original image of Tunisian city name "Methline"
(b) Suppression of diacritics
(c) The same word after the pre-processing steps and characteristic point's detection

An algorithm of segmentation of the skeleton makes the segments of a word. These segments are classified into three categories: (see figure 4)

- Segment 1: represents a stroke that is located between two end points or between an end point and a branching point (or crossing point).
- Segment 2: shows a stroke of link that is located between two branching points (or crossing) or between a branching point and a crossing point. This type of segment does not represent a contour of an occlusion.
- Segment 0: presents a stroke of link that is located between two branching points (or crossing) or between a branching point and a crossing point but it represents a contour of an occlusion.

The segmentation is released by an inspection of successive skeleton points. In an instance, one segment is limited between two characteristic points. As a result, this segmentation allows a first organization of the points of every segment. To facilitate this operation, we start with the extraction of the segments of type 1, type 2 and type 0. We eliminate the segments localized on the skeleton every time.

The temporal order reconstruction is made by the consideration of these criteria:

- Choose the direction right-left of the pixels displacement.
- Choose the minimum distance between pixels.
- Minimize the repetition of the segments.

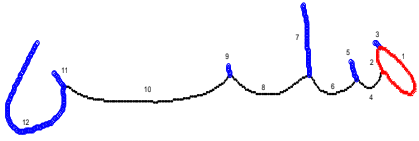

Figure 4. Segmentation of the skeleton of an Arabic word.

- Choose the minimum angular deviation in the crossing and branching points [27].

The sequence of these segments represents the original trajectory of the handwritten word. The rebuilt signal will be represented by a succession of X and Y coordinates.

The rebuilt signal does not contain the speed of the pen. A study made in the neuromuscular effect shows that the pen speed decreases at the beginning and the end of the stroke and in the angular variation of the curve. That is remarkable, if we observe an on-line signal acquired by a tablet. The on-line signal is acquired in real time. If we take into account the resolution (or the time of recording) of the tablet, it is noticed clearly that these points are not distributed in an equidistant manner. Concentrations of points are observed at the beginning and the end of the feature and in the curves of the stroke. This information is used in the on- line systems to calculate curvilinear speed. In order to obtain this information and to profit from on-line modelling of the writing, the rebuilt signal will be sampled by applying a method described in [9]. In order to modelling the rebuilt signal, we use the beta-elliptic representation developed by [18].

## V. BETA- ELLIPTIC APPROACH FOR HANDWRITING MODELLING

As it was explained in [17], the Beta-elliptic model considers a simple movement as the response to the neuromuscular system, which is described by an elliptic trajectory and a Beta velocity profile. Handwritten scripts are, then, segmented into simple movements, as already mentioned, called strokes, and are the result of a superimposition of time-overlapped velocity profiles. In our approach of modelling, a simple stroke is approximated by a Beta profile in the dynamic domain which corresponds in turn to an elliptic arc in the static domain. As it was explained in [17], [18], the complete velocity profile of the neuromuscular system will be described by a Beta model as follows:

$$\beta(t, q, p, t_0, t_1) = \begin{cases} \left(\frac{t-t_0}{t_c-t_0}\right)^p \left(\frac{t_1-t}{t_1-t_c}\right)^q & \text{if } t \in [t_0, t_1] \\ 0 & \text{elsewhere} \end{cases} \quad (1)$$

Where :

p and q are intermediate parameters, which have an influence on the symmetry and the width of Beta shape, $t_0$ is the starting time of Beta function, $t_c$ is the instant when the curvilinear velocity reaches the amplitude of the inflexion point, $t_1$ is the ending time of Beta function, $t_0 < t_1 \in IR$ (IR is the real set) and :

$$t_c = \frac{p \times t_1 + q \times t_0}{p+q} \quad (2)$$

We also consider the geometric representation of the handwritten trajectory. In the geometric plan, the trajectory is represented by a sequence of elliptic arcs [17]. The elliptic model is a static model.

The elliptic equation is written as follows:

$$\frac{X^2}{a^2} + \frac{Y^2}{b^2} = 1 \quad (3)$$

Consequently, a stroke is characterized by seven parameters. The first four Beta parameters (t0, t1, p and k) reflect the global timing properties of the neuromuscular networks involved in generating the movement (k=1), whereas the last three elliptic parameters (θ, a and b) describe the global geometric properties of the set of muscles and joints recruited to execute the movement, θ is the angle of elliptical stroke, a and b are respectively the big and small axes of the ellipse. The result of Beta model reconstruction of signal velocity is shown in figure 5.

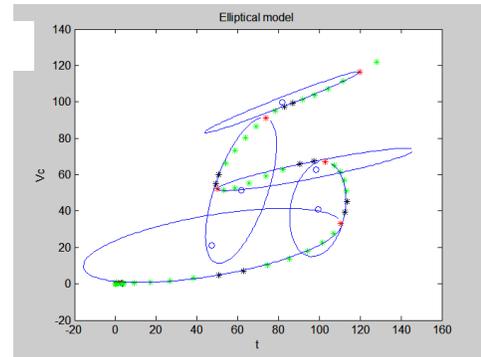

Figure 5. Example of elliptic representation of digit 5

## VI. RECOGNITION SYSTEM AND EXPERIMENTAL RESULTS

The graph matching algorithm performs a distance computation between two trajectories. A nearest neighbourhood algorithm is used to associate the nearest points between graph trajectories. The Euclidian distance is then calculated in order to evaluate the graphic similarity. Note, that no deformation is assumed to the graphs during the processing. Figure 6 shows a fragment of handwritten graphs that had been superposed and scaled, then associated points (see figures 6a and 6b). If N1 and N2 are respectively the number of strokes of graph 1 and graph 2, so the distance between the traces of strokes is calculated by the formula 4. Every stroke will be represented by a middle point [28].

$$Dist_{1,2} = \frac{1}{N}\sum_{i=1}^{N}(dist_{1,2})_i + P*(|N_1 - N_2|) \quad (4)$$

$$P = Max_{i=0}^{N}(dist_{1,2})_i \quad (5)$$

Where N= min (N$_1$, N$_2$), P is a penalty (formula 5)

and $(dist_{1,2})_i$ is the Euclidian distance between two associated points (formula 6).

$$(dist_{1,2})_i = \sqrt{(X_i^1 - X_i^2)^T * (X_i^1 - X_i^2)} \qquad (6)$$

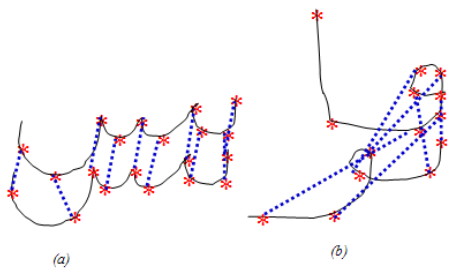

Figure 6. Graph matching process

(a) two similar graphs
(b) two different graphs

To test the developed system, we used the dataset of envelopes which have been created in our laboratory. It is made up of one thousand images of handwritten addresses, 740 of which are in Arabic and 260 in Latin symbols.

The addresses images are obtained via scanning envelopes collected from the education services files of two academic establishments in Sfax (see figure 7).

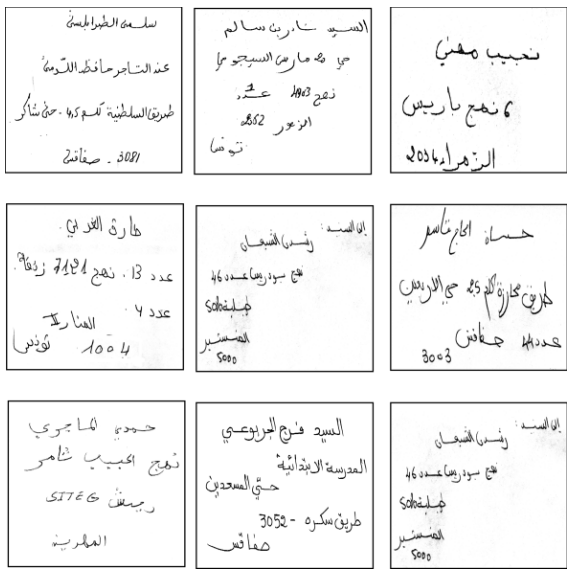

Figure 7. Some samples of images of addresses.

The scanning was done with 300 ppi resolution and 256 colours. The Tunisian postal code consists of 4 digits. Our data set of addresses consists of 1000 addresses enabling to have a data set of 4000 handwritten digits. The recognition system of the developed postal code address developed is divided into pre-processing steps and subsequent classifier. The processing steps were done by a serial mode which consists in filtering, smoothing, order reconstruction of the handwritten trajectory and the beta elliptic modelling representation.

The results are satisfactory and encouraging: 97% of the total envelopes images have been filtered and smoothed. 3% will be manually classified because of the existence of noises and imperfections in handwritten addresses.

In order to benefit of the on line feature, we reconstructed the temporal order of the handwritten words. Compared [1] to other systems, our modelling system based on neuro-physiological approach performs better. In fact, we reduced the vector size representing word by using the beta elliptic representation. To recognize the postal city name, we developed the graph matching algorithm. The recognition rate obtained is about 98%. This means that out of 100 envelopes, 98 are recognized and only 2 are released and will be processed manually.

## VII. CONCLUSION

The postal automation is one of the most active OCR applications. It becomes indispensable to ensure a fast postal service. These systems have drawn more and more interest. Compared to results obtained using neural network [5], our results are more promising. In face of the complexity and the variability of the handwritten words, the results obtained are acceptable and very promising.

## VIII. FUTURE WORKS

We purport to pursue our work to improve our system related to processing the handwritten digits and characters. We hope to reach the performances of international systems conceived for the automatic processing of postal addresses. Also, we intend to extend our study to the areas of many administrative forms and bank checks processing applications.

### ACKNOWLEDGMENTS

This work is partly supported by the project financed by the "Fond International de Coopération Universitaire" (FICU). The authors would like to address particular thanks to M. Cheriet, G. Stamon, J. Suen, E. Lecolinet for their help and contribution during all the phases of project realization. Also, the authors would like to acknowledge the financial support of this work by grants from the General Department of Scientific Research and Technological Renovation (DGRST), Tunisia, under the ARUB program 01/UR/11/02.

AUTHORS PROFILE

Moncef Charfi Profile

Moncef Charfi was born in Sfax (Tunisia), in 1950. He received his Ph.D. from the University Paris-Sud Orsay (France) in 1980. He is currently HDR in Computer Systems Engineering and Assistant Professor in the Department of Computer Engineering and Applied Mathematics at the National School of Engineers of Sfax University of Sfax,. Member in the Research Group on Intelligent Machine REGIM. He is interested in his research to the field of Document Analysis and Recognition and the processing of the old documents. He reviewed several scientific articles. He is IEEE member, and Arab Computer Society (ACS) member.

Monji Kherallah Profile

Monji Kherallah was born in Sfax, Tunisia, in 1963. He received the Engineer Diploma degree and the Ph.D both in electrical engineering, in 1989 and 2008, from University of Sfax (ENIS). For fourteen years ago, he was an engineer in Biotechnology Centre of Sfax. Now he teaching in Faculty of Science of Sfax and member in Research Group of Intelligent Machines: REGIM. His research interest includes the Handwritten Documents Analysis and Recognition. The techniques used are based on intelligent methods, such as neural network, fuzzy logic, genetic algorithm etc. He is one of the developers of the ADAB-Database (used by more than 50 research groups from more than 10 countries). He co-organized the Arabic Handwriting Recognition Competitions at the Online Arabic Handwriting Competitions at ICDAR 2009 and ICDAR 2011. He has more than 40 papers, including journal papers and book chapters. He is a member of IEEE and IEEE AESS Tunisia Chapter Chair, 2010 and 2011. He is reviewer of several international journals.

Abdelkarim El-Baati Profile

Abdelkarim ELBAATI was born in Chebba, Tunisia, in 1976. He received the Ph.D in electrical engineering, in 2009, from University of Sfax (ENIS). Now he teaching in Higher Institute of Applied Sciences and Technology of Mahdia and member in Research Group of Intelligent Machines: REGIM. His research interest includes the Handwritten Documents Analysis and Recognition. The techniques used are based on intelligent methods, such as HMM, genetic algorithm etc. He has more than 10 papers, including journal papers. He is a member of IEEE . He is reviewer of PRL journal.

Adel M. Alimi Profile

Adel M. Alimi was born in Sfax (Tunisia) in 1966. He graduated in Electrical Engineering 1990, obtained a PhD and then an HDR both in Electrical & Computer Engineering in 1995 and 2000 respectively. He is now professor in Electrical & Computer Engineering at the University of Sfax. His research interest includes applications of intelligent methods (neural networks, fuzzy logic, evolutionary algorithms) to pattern recognition, robotic systems, vision systems, and industrial processes. He focuses his research on intelligent pattern recognition, learning, analysis and intelligent control of large scale complex systems. He is associate editor and member of the editorial board of many international scientific journals (e.g. "IEEE Trans. Fuzzy Systems", "Pattern Recognition Letters", "NeuroComputing", "Neural Processing Letters", "International Journal of Image and Graphics", "Neural Computing and Applications", "International Journal of Robotics and Automation", "International Journal of Systems Science", etc.).

He was guest editor of several special issues of international journals (e.g. Fuzzy Sets & Systems, Soft Computing, Journal of Decision Systems, Integrated Computer Aided Engineering, Systems Analysis Modelling and Simulations).

He is the Founder and Chair of many IEEE Chapter in Tunisia section, he is IEEE Sfax Subsection Chair (2011), IEEE ENIS Student Branch Counsellor (2011), IEEE Systems, Man, and Cybernetics Society Tunisia Chapter Chair (2011), IEEE Computer Society Tunisia Chapter Chair (2011), he is also Expert evaluator for the European Agency for Research. He was the general chairman of the International Conference on Machine Intelligence ACIDCA-ICMI'2005 & 2000. He is an IEEE senior member.